\UseRawInputEncoding
\documentclass[letterpaper, 10 pt, conference]{ieeeconf}  % Comment this line out if you need a4paper

\usepackage{booktabs}
\usepackage{amssymb}
\usepackage{amsmath}
\usepackage{graphicx}
\usepackage{tabularx}
\usepackage{array}
\usepackage[table]{xcolor}

\IEEEoverridecommandlockouts                              % This command is only needed if 
\newcommand{\up}[1]{%
\textcolor{red}{\scriptsize\,$\uparrow$#1}}
\newcommand{\down}[1]{%
\textcolor{blue}{\scriptsize\,$\downarrow$#1}}
\title{\LARGE \bf
SocialCoAct: Enabling and Benchmarking Egocentric Social Co-Action for VLM-Based Humanoid Robots
}

\author{Jiancong Ma$^{1}$ \quad Kun Li$^{1}$ \quad Jin Jiang$^{1}$ \quad Shengcai Liao$^{1*}$
\thanks{$^{1}$Jiancong Ma, Kun Li, Jin Jiang and Shengcai Liao are with the College
of Computing and Artificial Intelligence (CCAI), United Arab Emirates
University (UAEU), Al Ain, Abu Dhabi, United Arab Emirates.}% <-this % stops a space
\thanks{$^*$Corresponding author: Shengcai Liao (email: scliao@ieee.org).}}%

\usepackage{makecell}

\begin{document}

\maketitle
\thispagestyle{empty}
\pagestyle{empty}

\newcommand{\frameworkname}{SocialCoAct}
\newcommand{\agentname}{SocialCoAct-Agent}
\newcommand{\benchname}{SocialCoAct-Bench}

\title{\LARGE \bf
SocialCoAct: Enabling and Benchmarking Egocentric Social Co-Action for VLM-Based Humanoid Robots
}

% \author{Jiancong Ma$^{1}$ and Shengcai Liao$^{2}$% <-this % stops a space
% \thanks{*This work was not supported by any organization}% <-this % stops a space
% \thanks{$^{1}$Albert Author is with Faculty of Electrical Engineering, Mathematics and Computer Science,
%         University of Twente, 7500 AE Enschede, The Netherlands
%         {\tt\small albert.author@papercept.net}}%
% \thanks{$^{2}$Bernard D. Researcheris with the Department of Electrical Engineering, Wright State University,
%         Dayton, OH 45435, USA
%         {\tt\small b.d.researcher@ieee.org}}%
% }

%%%%%%%%%%%%%%%%%%%%%%%%%%%%%%%%%%%%%%%%%%%%%%%%%%%%%%%%%%%%%%%%%%%%%%%%%%%%%%%%
\begin{abstract}
Enabling humanoid robots to participate effectively in social interactions is important for their integration into human-centered environments. Existing methods, however, typically map pre-generated speech directly to continuous motion trajectories but have limited awareness of complex social environments and lack explicit high-level semantic action planning. We introduce SocialCoAct, a unified agent-centered framework for VLM-based egocentric social co-action. We formulate egocentric social co-action as an end-to-end agent task and develop SocialCoAct-Agent, which transforms egocentric observations and dialogue history into verbal responses and atomic action sequences, aligns each action with its corresponding utterance span, and executes the resulting behaviors in MuJoCo and on a physical humanoid robot. To evaluate the VLM capabilities required by this task, we construct SocialCoAct-Bench from continuous egocentric recordings of natural multi-party interactions. The benchmark comprises 423 human-verified multiple-choice questions across three subtasks: Long Episode Understanding (LEU), Speaker Intention Understanding (SIU), and Coordinated Action Planning (CAP). Across a diverse set of closed-source and open-weight VLMs, including general and egocentric models, the highest overall accuracy is 77.54\%, while accuracy on the hard subset reaches only 46.51\%. Extended reasoning improves Qwen3-VL-2B and Qwen3-VL-4B by 19.62 and 17.02 percentage points, respectively, but reduces inference throughput by 53.6$\times$ and 20.2$\times$. Visual-input ablations further show that the benefits of visual context vary across models. SocialCoAct provides a foundation for developing and evaluating VLM-based humanoid agents that coordinate context-aware speech and action in complex social interactions. The codebase and benchmark will be made publicly available upon publication.
\end{abstract}

\section{Introduction}
\label{sec:introduction}
Humanoid robots are increasingly expected to participate in everyday social interactions. An ideal humanoid robot should be able to visually comprehend complex social scenes and ongoing activities, infer the intentions conveyed through human language, and respond naturally through coordinated verbal and bodily behaviors~\cite{Payandeh2025SocialLLaVA,Vinanzi2025CASPER}. Such responses should not only be semantically appropriate but also align with the interaction context, timing, and intended addressee, thereby resembling the integrated way in which humans communicate through speech and action~\cite{Galatolo2025TextGesture,Beckova2025MultimodalExplainability}.

Recent systems can generate synchronized co-speech gestures for physical robots, while a few can jointly produce verbal and embodied responses~\cite{cospeechli2024}. However, their gestures are typically conditioned on pre-generated text or speech rather than grounded in an online visual understanding of evolving social interactions~\cite{Oralbayeva2024CommunicativeBehaviour}. They may also inherit idiosyncratic styles from individual demonstrators without explicitly modeling semantic alignment. Consequently, existing evaluations do not jointly assess whether verbal and bodily responses are appropriate given scene dynamics, conversational history, speaker intention, ongoing activities, and addressee relations in multi-party interactions.

Evaluating this capability poses a parallel challenge. Social appropriateness cannot be judged from an isolated frame or utterance; it depends on conversational history, speaker intention, turn-taking dynamics, addressee relations, and the norms of the ongoing activity. Moreover, because humans communicate through coordinated speech and bodily behavior, an evaluation must consider not only the appropriateness of each modality but also their semantic alignment~\cite{Oralbayeva2024CommunicativeBehaviour}. Existing evaluations of robotic social intelligence, however, primarily focus on human--robot interaction or the recognition and correction of observed social behaviors~\cite{LeeD-RSS-26}. Although these tasks assess the perception of social cues and inappropriate behavior, they rarely examine whether a humanoid agent can actively generate a contextually appropriate verbal response together with semantically aligned bodily actions.

% \mjc{You are required to explain that reliable response of a dialogue comes from the detailed long context(what the specific scene,same words means different in different scenes. Understanding the link between actions and detailed scene.)}

\begin{figure*}[t!]
\centering
\includegraphics[width=\linewidth]{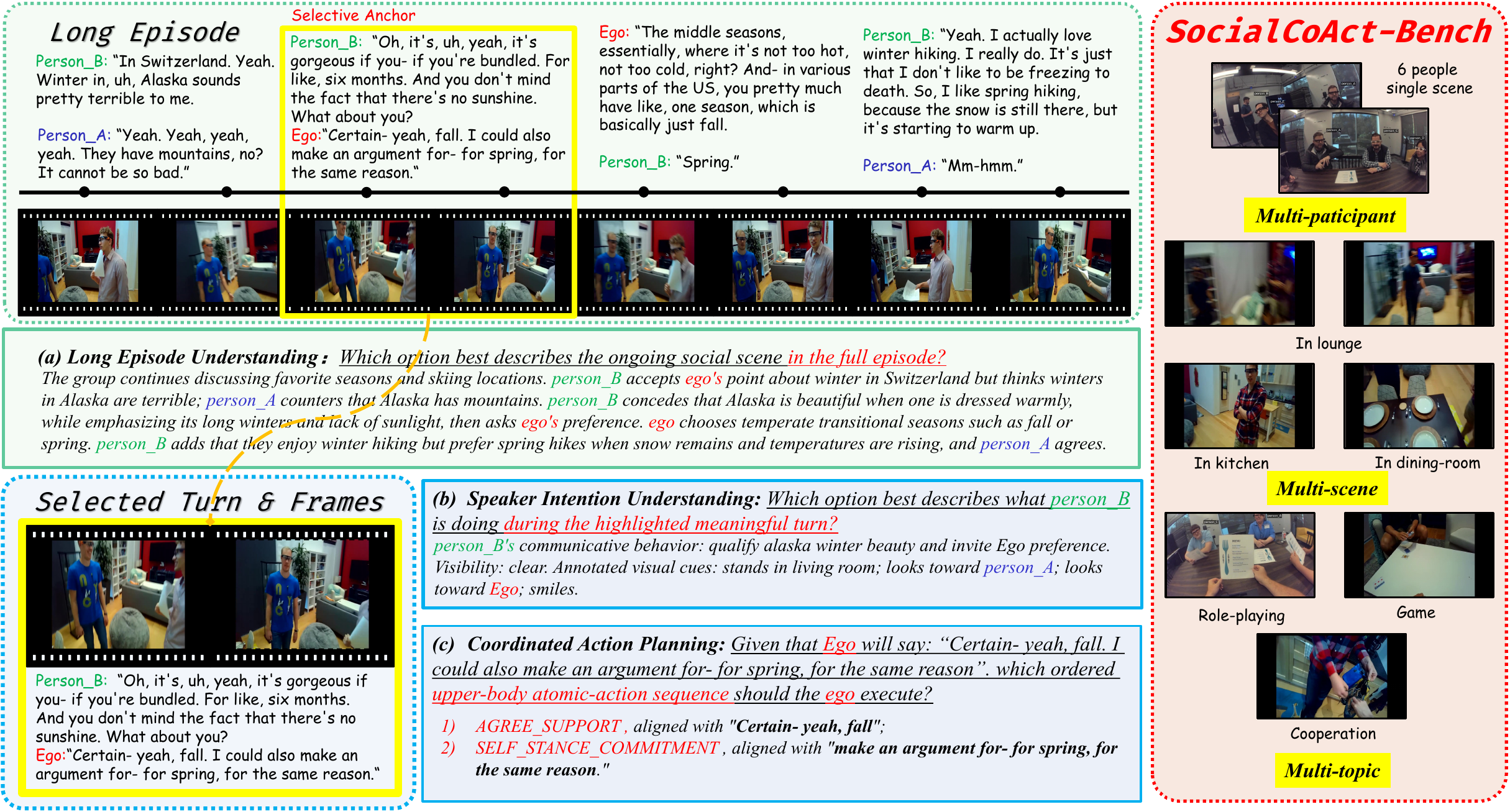}
\caption{Overview of SocialCoAct-Bench, which comprises three subtasks: Long Episode Understanding (LEU), Speaker Intention Understanding (SIU), and Coordinated Action Planning (CAP).}
\label{fig:oview}
\end{figure*}

To address this gap, we introduce \textbf{\frameworkname{}} (\textbf{Social} \textbf{Co}ordinated \textbf{Act}ion), a unified VLM-based framework for egocentric social co-action in humanoid robots. The framework includes \textbf{\agentname{}}, which reasons about the ongoing interaction and the current speaker's intent from egocentric observations and dialogue history, generates a verbal response together with an ordered sequence of atomic actions, aligns each action with its corresponding utterance span, and executes the resulting co-speech behavior in MuJoCo and on a physical humanoid robot. To evaluate whether current VLMs can support this task, we construct \textbf{\benchname{}} from continuous first-person recordings of diverse, natural multi-party interactions.
As shown in Fig.~\ref{fig:oview}, SocialCoAct-Bench evaluates the VLM capabilities required by this agent task at two complementary temporal levels through three subtasks. At the long-context level, \textbf{Long Episode Understanding (LEU)} assesses the model's understanding of the overall activity and interaction context across an egocentric episode. At the short-turn level, \textbf{Speaker Intention Understanding (SIU)} evaluates the interpretation of the current speaker's intention, while \textbf{Coordinated Action Planning (CAP)} assesses whether and how the agent should produce an ordered plan of predefined atomic actions aligned with spans in a given verbal reply.

Our contributions can be summarized as follows:
\begin{enumerate}
\item We develop \textbf{\agentname{}}, an agent capable of transforming egocentric observations and dialogue history into verbal responses and utterance-aligned atomic actions executable in MuJoCo and on a physical humanoid robot.

\item To evaluate the capabilities required by this agent, we introduce \textbf{SocialCoAct-Bench}, an egocentric benchmark of 423 human-verified questions organized into three progressive tasks: \textbf{LEU} for long-context understanding, \textbf{SIU} for speaker-intention understanding, and \textbf{CAP} for planning atomic actions aligned with a given verbal reply in complex multi-party interactions.

\item We benchmark a diverse set of closed-source and open-weight VLMs, including general and egocentric models, and characterize their capability boundaries across tasks and difficulty levels. Our reasoning ablations demonstrate substantial trade-offs between accuracy and efficiency, while our modality ablations show that the benefits of visual context vary across models.
\end{enumerate}
%%%%%%%%%%%%%%%%%%%%%%%%%%%%%%%%%%%%%%%%%%%%%%%%%%%%%%%%%%%%%%%%%%%%%%%%%%%%%%%%
\section{RELATED WORK}
\subsection{Embodied Social Response Generation}
Existing research on embodied social response generation spans facial reaction
generation, gesture synthesis, and multimodal conversational behavior.
Responsive Listening Head Generation introduces a benchmark and baseline for
synthesizing listener head movements and facial expressions from a speaker's
audiovisual behavior, while REACT defines online and offline settings for
generating diverse and appropriate facial reactions in dyadic
conversations~\cite{zhou2022responsive,song2024react}. More recent systems capture richer interaction
dynamics. PolySLGen generates speaking and listening reactions for a target
participant based on the speech and motion history of polyadic
interactions~\cite{lin2026polyslgen}, whereas ProAct combines long-horizon social reasoning
with streaming nonverbal behavior to enable proactive embodied
interaction~\cite{zhang2026proact}.

Collectively, these studies have advanced response-generation benchmarks,
generative architectures, and integrated interactive systems. However, their
evaluations are generally not designed to isolate the upstream decision-making
capabilities required of VLMs, including long-horizon contextual understanding,
speaker-intent inference, and the construction of language-aligned action plans
for humanoid robots.

\begin{figure*}[t!]
\centering
\includegraphics[width=\linewidth]{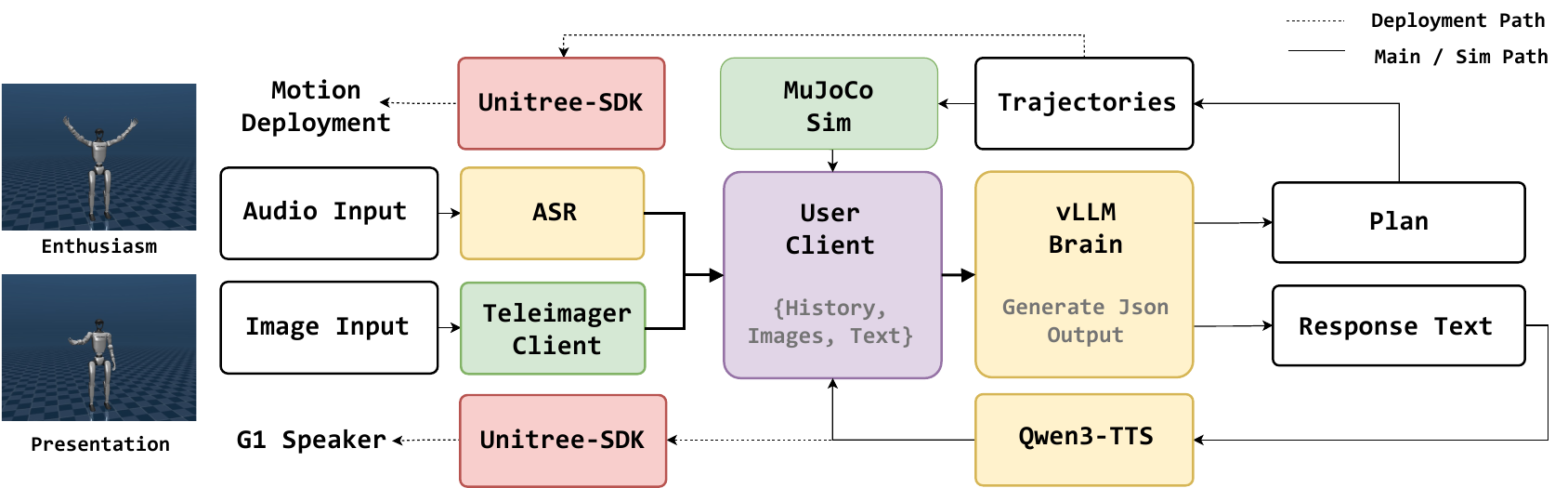}
\caption{Overview of the SocialCoAct-Agent architecture. The leftmost panels show examples of atomic robot actions in MuJoCo.}
\label{fig:agent}
\end{figure*}

\subsection{Multimodal Social and Egocentric Benchmarks}
Existing benchmarks evaluate different aspects of social intelligence. Social-IQ and Social Genome assess multimodal reasoning about social situations, whereas SIV-Bench further evaluates social scene understanding, latent-state reasoning, and interaction dynamics prediction. SocialGesture focuses on gesture recognition, temporal localization, and vision-language reasoning in multi-person interactions~\cite{zadeh2019socialiq,mathur2025socialgenome,kong2026siv,cao2025socialgesture}. EasyCom, EgoCom, and Ego4D provide first-person recordings of natural communication, whereas JRDB-Social captures individual behaviors and group dynamics from the perspective of a mobile robot~\cite{donley2021easycom,northcutt2023egocom,grauman2022ego4d,jahangard2024jrdbsocial}.

Recent benchmarks have increasingly emphasized interactive decision-making. HoloAssist evaluates mistake detection and intervention-type prediction in egocentric task assistance, whereas OmniMMI benchmarks proactive reasoning and turn-taking in streaming multimodal contexts~\cite{Wang_2023_ICCV,Wang_2025_CVPR}. In embodied human--robot interaction (HRI), SHREC evaluates social errors, competencies, interaction flow, and corrective actions using videos of real-world human--robot interactions, whereas ERR@HRI 2.0 focuses on multimodal failure detection in dyadic human--robot conversations~\cite{LeeD-RSS-26,cao2025errhri}.

Nevertheless, these benchmarks largely evaluate social perception, reasoning, error diagnosis, and response generation in isolation. They do not jointly assess the VLM capabilities needed to understand an evolving multi-party interaction and plan bodily actions that are semantically aligned with a given verbal response.

\section{SocialCoAct}
\subsection{SocialCoAct-Agent Architecture}
In this section, we present the architecture of \agentname{}. The agent is built on the vLLM codebase~\cite{kwon2023efficient}. Fig.~\ref{fig:agent} illustrates its main components and overall workflow.
\begin{figure*}[t!]
\centering
\includegraphics[width=\linewidth]{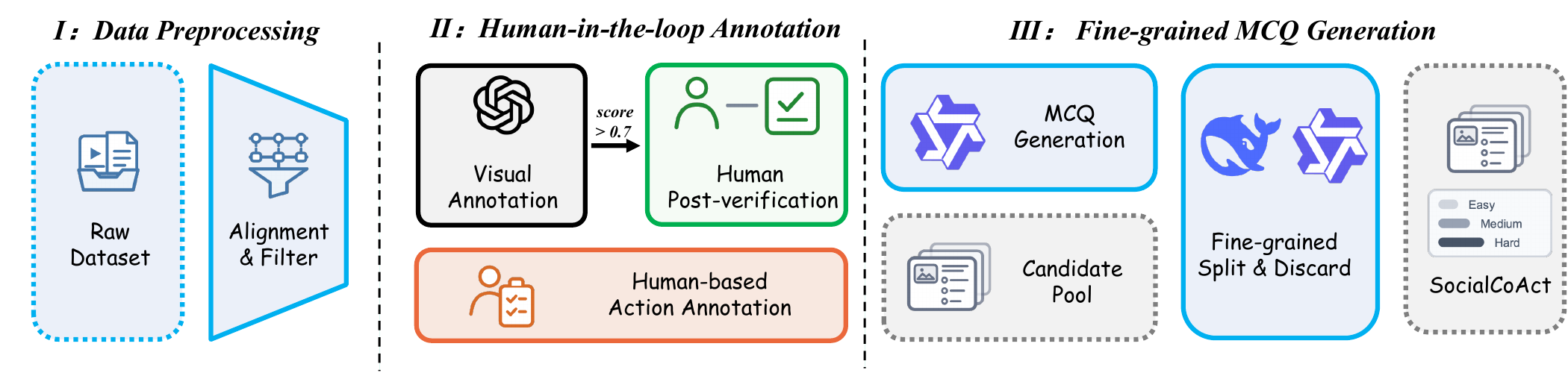}
\caption{Overview of the SocialCoAct-Bench construction pipeline. 
% Dashed boxes indicate data at different stages, while solid boxes indicate pipeline components.
}
\label{fig:pipe}
\end{figure*}
At each interaction turn, \agentname{} receives both speech and visual observations. The user's speech is recorded through the microphone as an audio input $a_t$ and transcribed by the Whisper-based ASR service~\cite{radford2023robust}:
\[x_t = f_{\mathrm{ASR}}(a_t),
\]

% \begin{equation}
% \label{eq:1}
%     x_t = f_{\mathrm{ASR}}(a_t),
% \end{equation}
where $x_t$ denotes the transcribed instruction. In parallel, the head-mounted camera continuously streams images from the robot PC through Teleimager. The client retrieves the most recent RGB frames $I_t$ when an interaction request is issued. The client then combines the instruction $x_t$, the images $I_t$, and the conversation history $H_t$ into a multimodal planning request with:
% \begin{equation}
% \label{eq:2}
%     R_t = \{x_t, I_t, H_t\}.
% \end{equation}
\[
R_t = \{x_t, I_t, H_t\}.
\]
The request is processed by \agentname{}. The model jointly interprets the textual instruction, visual observation, and dialogue context and produces a structured response. The output parser separates the model output into a natural-language response $s_t$, an ordered execution sequence $E_t$, and a set of atomic action plans $P_t$:
% \begin{equation}
%     (s_t, E_t, P_t) = f_{\theta}(x_t, I_t, H_t).
% \end{equation}
\[
(s_t, E_t, P_t)
= f_{\theta}(R_t)
= f_{\theta}(x_t, I_t, H_t).
\]
Here, $E_t$ preserves the ordering between speech segments and their associated actions, while each element of $P_t$ specifies an atomic action and its semantic anchor. The agent does not directly execute joint values generated by the language model. Instead, each predicted action is validated against a predefined action whitelist and mapped to a corresponding trajectory. The resulting response, containing \texttt{response\_text} and \texttt{plans}, is returned to the client as a structured JSON message.

After receiving the planning result, the client dispatches the response to separate speech and motion branches. In the speech branch, the natural-language response is synthesized by Qwen3-TTS~\cite{hu2026qwen3}:
% \begin{equation}
%     \hat{a}_t = f_{\mathrm{TTS}}(s_t),
% \end{equation}
\[
\hat{a}_t = f_{\mathrm{TTS}}(s_t),
\]
where $\hat{a}_t$ is the generated audio returned to the interface for playback. In the motion branch, the ordered action plans are converted into predefined joint trajectories. The lower body remains fixed in the standing configuration, while the atomic trajectories are processed using interpolation, joint-velocity limiting, and smooth transition functions. In the current simulation setting, the resulting trajectory is rendered in MuJoCo and returned to the interface. The same structured action plan can optionally be forwarded to the robot-side controller for physical execution.

The semantic anchors in $P_t$ maintain a turn-level correspondence between speech content and gestures. The current implementation generates the complete planning result before dispatching speech synthesis and action execution.
\subsection{Benchmark Construction Pipeline}

This section outlines the construction pipeline of SocialCoAct-Bench. The process comprises three progressive stages, each corresponding to a subsection: \textit{Data Preprocessing} (\textit{Sec.}~\ref{sec:data-preprocessing}), \textit{Human-in-the-loop Labeling} (\textit{Sec.}~\ref{sec:HIL_label}), and \textit{Fine-grained MCQ Generation} (\textit{Sec.}~\ref{sec:mcq-generation}). Fig.~\ref{fig:pipe} summarizes the construction pipeline.
\subsubsection{Data Preprocessing}
\label{sec:data-preprocessing}
SocialCoAct-Bench is constructed from two public egocentric interaction datasets, EasyCom~\cite{donley2021easycom} and EgoCom~\cite{northcutt2023egocom}. We temporally align each video with its transcript and apply semantic filtering to identify meaningful conversational turns. Tag-only events, unintelligible fragments, fillers, and isolated low-information acknowledgments are excluded as annotation targets. Episode durations are restricted to 12--60 seconds. All videos are standardized to a resolution of $1280 \times 720$ at 10 FPS and paired with timestamped transcripts.

\subsubsection{Human-in-the-loop Labeling}
\label{sec:HIL_label}
Given the distinct characteristics of these tasks, we adopt a hybrid annotation strategy. The visual understanding tasks are annotated through model-assisted pre-labeling followed by expert validation, whereas the action-planning task currently relies more heavily on human judgment because of its subjective and context-sensitive nature.
We first select representative turns and episodes that are suitable for annotation. For each selected video segment, we employ GPT-5.6~\cite{openai2026gpt56} to generate initial annotations, including a high-level description of the ongoing activity and an inference of the current speaker's intention and expression. Each prediction is accompanied by a confidence score $c$, and predictions with $c < \tau$, where $\tau = 0.7$, are discarded. Domain experts then independently review the remaining pre-labeled outputs to verify their factual correctness and semantic appropriateness.

To annotate appropriate humanoid actions, we rely exclusively on human annotators. These annotations involve inherently pragmatic and subjective judgments that require a nuanced understanding of social context, conversational dynamics, and cultural norms. Each short conversational turn is evaluated by three human annotators, who select the most appropriate expressive action from a predefined vocabulary of 23 atomic action types. Each action type has a distinct communicative purpose and affective meaning, such as whether an action conveys a positive or negative attitude or serves to inquire, acknowledge, encourage, reject, or express another intention. When disagreements arise, the final annotation is determined by majority vote among the three annotators. To promote cross-cultural recognizability and broad acceptability, the action vocabulary was co-designed with participants from diverse cultural backgrounds to minimize conflicting interpretations across cultural contexts. Importantly, the annotated actions do not necessarily reproduce the ego agent's observed real-world behavior. Instead, they represent broadly understandable expressive signals that communicate the intended social meaning within the benchmark setting.
\subsubsection{Fine-grained MCQ Generation}
\label{sec:mcq-generation}

Based on the human-verified annotations, we construct three task-specific sets of four-option multiple-choice questions (MCQs): LEU, SIU, and CAP\@. For each sample, the question and correct answer are derived directly from the corresponding task annotation. We use Qwen3-VL-8B~\cite{bai2025qwen3vl} only to generate distractor options that remain contextually relevant but contain identifiable semantic errors with respect to the ground-truth annotation. Such errors include, for example, reversing the expressed sentiment from positive to negative or replacing an inquiry with a clearly different communicative intention.

\begin{table*}[t!]
\centering
\caption{Comparison with representative benchmarks from social, HRI, and egocentric domains. Abbreviations: I, V, A, T = image, video, audio, text; SS = social scene; ESR = embodied social response; AUA = action--utterance alignment.}
\label{tab:benchmark_comparison}
\setlength{\tabcolsep}{3.5pt}
\renewcommand{\arraystretch}{1.06}
\small
\begin{tabular*}{\textwidth}{
@{\extracolsep{\fill}} l c c c c c c @{}
}
\toprule
Benchmark
& Venue
& Egocentric
& Modalities
& \shortstack{SS}
& \shortstack{ESR}
& \shortstack{AUA} \\
\midrule
\multicolumn{7}{@{}l}{\textit{Social understanding benchmarks}} \\
Social-IQ~\cite{zadeh2019socialiq}
& CVPR'19
& $\times$
& V+A+T
& $\checkmark$
& $\times$
& $\times$ \\

Social Genome~\cite{mathur2025socialgenome}
& EMNLP'25
& $\times$
& V+A+T
& $\checkmark$
& $\times$
& $\times$ \\

SIV-Bench~\cite{kong2026siv}
& ACL'26
& $\times$
& V+A+T
& $\checkmark$
& $\times$
& $\times$ \\

\midrule
\multicolumn{7}{@{}l}{\textit{Human--robot interaction benchmarks}} \\

SCAND~\cite{scand2022ral}
& RA-L'22
& $\checkmark$
& V
& $\checkmark$
& $\times$
& $\times$ \\

JRDB-Social~\cite{jahangard2024jrdbsocial}
& CVPR'24
& $\checkmark$
& V+A
& $\checkmark$
& $\times$
& $\times$ \\

HRIBench~\cite{shi2025hribench}
& ISER'25
& $\checkmark$
& I+T
& $\checkmark$
& $\times$
& $\times$ \\

SocialNav-SUB~\cite{jesocialnav2025}
& CoRL'25
& $\checkmark$
& I+T
& $\checkmark$
& $\checkmark$
& $\times$ \\

SHREC~\cite{LeeD-RSS-26}
& RSS'26
& $\times$
& V+A+T
& $\checkmark$
& $\checkmark$
& $\times$ \\
\midrule
\multicolumn{7}{@{}l}{\textit{Egocentric benchmarks}} \\

Ego4D~\cite{grauman2022ego4d}
& CVPR'22
& $\checkmark$
& V+A+T
& $\checkmark$
& $\times$
& $\times$ \\

EgoTaskQA~\cite{egotaskqa2022}
& NeurIPS'22
& $\checkmark$
& V+T
& $\times$
& $\times$
& $\times$ \\

HoloAssist~\cite{Wang_2023_ICCV}
& ICCV'23
& $\checkmark$
& V+A+T
& $\times$
& $\times$
& $\times$ \\
EgoPlan-Bench~\cite{chen2026egoplan}
& IJCV'26
& $\checkmark$
& V+I+T
& $\times$
& $\times$
& $\times$ \\

EgoLife~\cite{yang2025egolife}
& CVPR'25
& $\checkmark$
& V+A+T
& $\checkmark$
& $\times$
& $\times$ \\

\textbf{SocialCoAct-Bench (Ours)}
& \textbf{--}
& \textbf{$\checkmark$}
& $\mathbf{V+T}$
& $\boldsymbol{\checkmark}$
& $\boldsymbol{\checkmark}$
& $\boldsymbol{\checkmark}$ \\

\bottomrule
\end{tabular*}
\vspace{1mm}
\end{table*}

All generated distractors are manually reviewed alongside their corresponding questions and correct answers. This review ensures that each MCQ is factually grounded in the episode and has a single, unambiguous correct answer. It also verifies that each distractor is sufficiently related to the episode context to be meaningful while remaining clearly incorrect because of a specific and identifiable discrepancy with the annotation. Distractors that lack an explicit error, introduce unsupported information, or create ambiguity are revised or removed.
After manual verification, each MCQ is independently answered by four expert models drawn from the DeepSeek and Qwen model families~\cite{deepseekai2026v4flashvisionexp}. We use the number of expert models that answer each question correctly as an empirical indicator of its difficulty. Specifically, questions answered correctly by at least three models are categorized as \textit{Easy}, those answered correctly by two models as \textit{Medium}, and those answered correctly by at most one model as \textit{Hard}. This procedure yields difficulty-stratified subsets for the final evaluation suite.

\begin{figure}
\centering
\includegraphics[width=\linewidth]{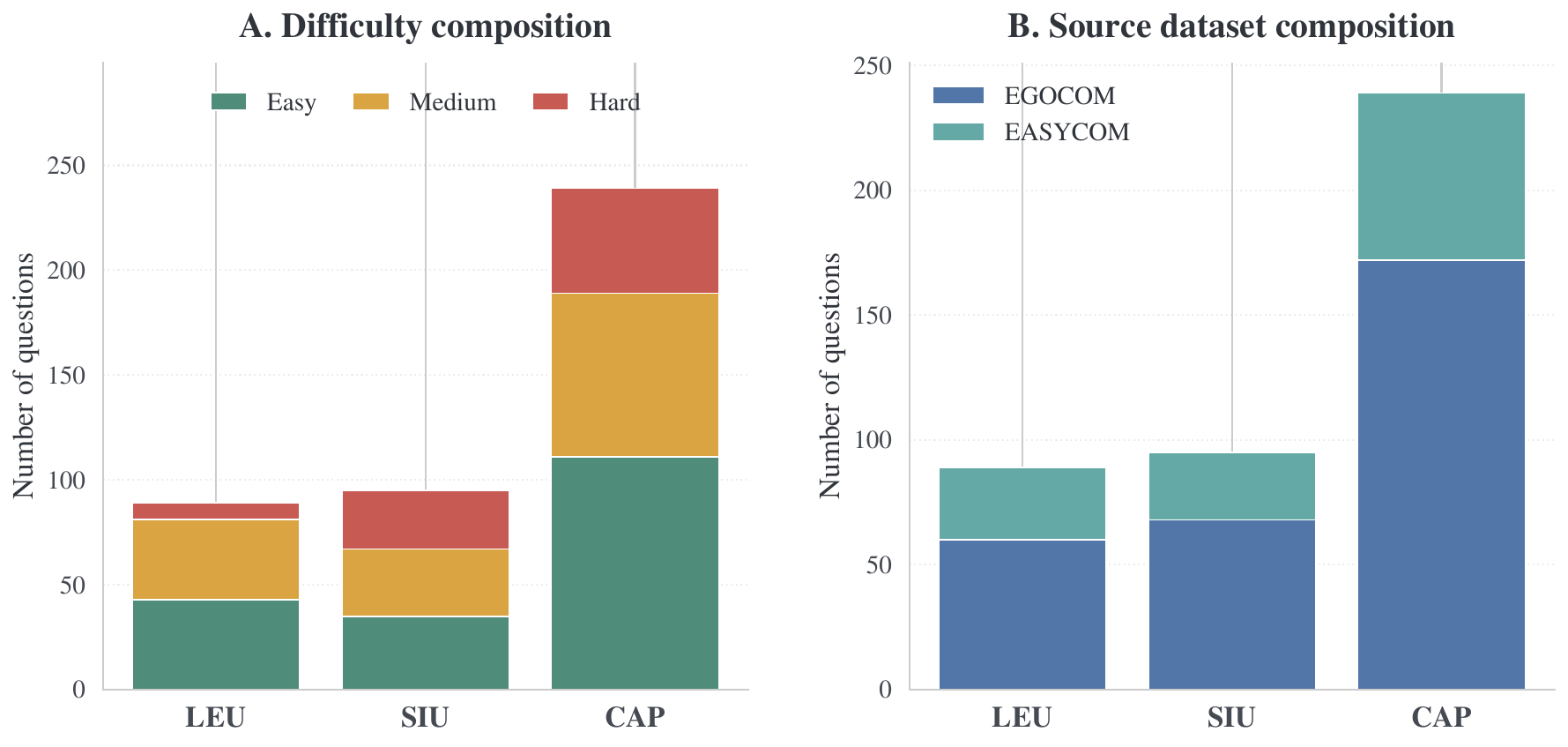}
% \caption{The overview of the filtered SocialCoAct benchmark.}
\caption{Difficulty and source distributions of SocialCoAct-Bench.}
\label{fig:stat}
\end{figure}

\subsection{Comparison with Other Benchmarks}
\label{sec:comparison}
In this section, we compare SocialCoAct-Bench with representative benchmarks in social understanding,
human--robot interaction (HRI), and egocentric domains. As summarized in
Table~\ref{tab:benchmark_comparison}, the comparison considers five
properties: whether the visual data are egocentric, the available modalities,
whether social interaction is an explicit evaluation target, whether the
model must determine an embodied social response, and whether individual
social actions are aligned with their corresponding utterance spans.

Existing social-understanding benchmarks primarily evaluate the interpretation of human behavior and communicative intent, while HRI benchmarks focus on perception, navigation, or the diagnosis and correction of robot social errors. Egocentric benchmarks provide first-person observations but mainly address activity recognition, long-term memory, or task-oriented action planning. SocialCoAct-Bench connects these directions by jointly
evaluating social understanding and embodied response planning from
multimodal egocentric observations. In particular, the CAP task represents each response as an ordered sequence of human-interpretable atomic social actions, with every action explicitly grounded in its corresponding utterance span. Table~\ref{tab:benchmark_comparison} shows that, among the benchmarks considered, SocialCoAct-Bench uniquely combines egocentric multimodal social understanding, embodied social response selection, and explicit alignment.

\section{Experiments}
\label{sec:experimental_setup}
\subsection{Models and Implementation}
The evaluation pipeline was implemented using VLMEvalKit~\cite{duan2024vlmevalkit}. Local inference, where applicable, was conducted on a single NVIDIA GeForce RTX 5090 GPU, while API-hosted models were accessed through their respective endpoints. We categorize the evaluated models according to weight availability and domain specialization. The first group comprises closed-source models: Gemini-3.1-Pro, Gemini-Flash-3.8, Claude-Sonnet-5, Grok-4.6 and GPT-5.6-Luna.~\cite{googledeepmind2026gemini31pro,googledeepmind2026gemini38flash,anthropic2026claudesonnet5,xai2026grok46,openai2026gpt56}. The second group consists of general open-weight VLMs: Qwen3.8-Max~\cite{qwen2026qwen38max}, DeepSeek-V4-Flash-Vision-Exp~\cite{deepseekai2026v4flashvisionexp}, InternVL2.5-1B/2B~\cite{chen2024internvl25}, Qwen3-VL-2B/4B~\cite{bai2025qwen3vl}, and Video-LLaVA-7B~\cite{lin-etal-2024-video}. The third group contains egocentric-domain open-weight VLMs: EgoVLM-360M, EgoVLM-3B~\cite{vinod2025egovlm}, EgoMind-7B~\cite{chenegomind2026}, and EgoThinker-v1~\cite{pei2026egothinker}.
\footnote{$^{\dagger}$ Although DeepSeek-V4-Flash-Vision-Exp and Qwen3.8-Max are open-weight models, we access them through the official API because of local hardware constraints.

$^{\ddagger}$ EgoVLM-360M processes one sampled frame per inference call; we aggregate the predictions by majority voting.}

\begin{table*}[t!]
\centering
\caption{
Performance comparison on SocialCoAct-Bench.
All values are accuracy (\%).
The best and second-best results in each column are highlighted in
\textbf{bold} and \underline{underlined}, respectively.}
\label{tab:main_results}
\setlength{\tabcolsep}{3.5pt}
\renewcommand{\arraystretch}{1.10}
\small
\resizebox{\textwidth}{!}{
%\begin{tabular}{lcccccccc}
\begin{tabularx}{\linewidth}{@{}>{\raggedright\arraybackslash}p{5cm} *{8}{>{\centering\arraybackslash}X}@{}}
\toprule
& & \multicolumn{1}{c}{Overall}
& \multicolumn{3}{c}{Task}
& \multicolumn{3}{c}{Difficulty} \\
\cmidrule(lr){3-3}
\cmidrule(lr){4-6}
\cmidrule(lr){7-9}
Model
& Params
& All
& CAP
& SIU
& LEU
& Easy
& Medium
& Hard \\
\midrule
Random Guess & -- & 25.00 & 25.00 & 25.00 & 25.00 & 25.00 & 25.00 & 25.00 \\
\midrule
\multicolumn{9}{l}{\textit{Closed-source models}} \\

Gemini-3.1-Pro~\cite{googledeepmind2026gemini31pro}
& --
& 69.74
& 56.90
& \underline{81.05}
& \underline{92.13}
& 78.31
& \underline{77.70}
& 37.21 \\

Claude-Sonnet-5~\cite{anthropic2026claudesonnet5}
& --
& \textbf{77.54}
& 71.55
& 80.00
& 91.01
& \textbf{92.59}
& 76.35
& \textbf{46.51} \\

Grok-4.6~\cite{xai2026grok46}
& --
& 47.52
& 53.14
& 47.37
& 32.58
& 59.26
& 42.57
& 30.23 \\

Gemini-Flash-3.8~\cite{googledeepmind2026gemini38flash}
& --
& 72.10
& 63.18
& \textbf{83.16}
& 84.27
& 85.71
& 74.32
& 38.37 \\

GPT-5.6-Luna~\cite{openai2026gpt56}
& --
& \underline{76.12}
& \textbf{72.80}
& \underline{81.05}
& 79.78
& \underline{87.83}
& \textbf{81.76}
& \underline{40.70} \\

\midrule
\multicolumn{9}{l}{\textit{General open-weight VLMs}} \\

InternVL2.5-1B~\cite{chen2024internvl25}
& 1B
& 36.64
& 39.33
& 40.00
& 25.84
& 56.08
& 18.92
& 24.42 \\

InternVL2.5-2B~\cite{chen2024internvl25}
& 2B
& 31.68
& 26.78
& 41.05
& 34.83
& 37.04
& 29.05
& 24.42 \\

Qwen3-VL-2B-Instruct~\cite{bai2025qwen3vl}
& 2B
& 39.48
& 45.19
& 22.11
& 42.70
& 64.02
& 20.27
& 18.60 \\

Qwen3-VL-4B-Instruct~\cite{bai2025qwen3vl}
& 4B
& 56.03
& 52.30
& 53.68
& 68.54
& 67.72
& 52.03
& 37.21 \\

Video-LLaVA-7B~\cite{lin-etal-2024-video}
& 7B
& 27.42
& 30.54
& 23.16
& 23.60
& 22.75
& 33.11
& 27.91 \\

DeepSeek-V4-Flash-Vision-Exp~\cite{deepseekai2026v4flashvisionexp}$\dagger$
& 305B
& 73.52
& \underline{72.38}
& 62.11
& 88.76
% & \textbf{98.41}
% & \underline{82.43}
% & 3.49 \\
& --
& --
& -- \\
Qwen3.8-Max~\cite{qwen2026qwen38max}$\dagger$
& 2.4T
& \textbf{77.54}
& 69.87
& \underline{81.05}
& \textbf{94.38}
% & \underline{--}
% & \textbf{--}
& --
& --
& -- \\
\midrule
\multicolumn{9}{l}{\textit{Egocentric-domain open-weight VLMs}} \\

EgoVLM-360M~\cite{vinod2025egovlm} $\ddagger$
& 0.36B
& 26.48
& 24.69
& 32.63
& 24.72
& 29.10
& 26.35
& 20.93 \\

EgoVLM-3B~\cite{vinod2025egovlm}
& 3B
& 38.77
& 21.76
& 58.95
& 62.92
& 49.74
& 33.11
& 24.42 \\

EgoMind-7B~\cite{chenegomind2026}
& 8.3B
& 53.19
& 48.54
& 58.95
& 59.55
& 68.25
& 48.65
& 27.91 \\

EgoThinker-v1~\cite{pei2026egothinker}
& 8.3B
& 53.43
& 50.63
& 63.16
& 50.56
& 69.84
& 42.57
& 36.05 \\

\bottomrule
\end{tabularx}}
\end{table*}

\subsection{Inference and Evaluation}
SocialCoAct-Bench comprises 423 multiple-choice questions: 239 for CAP, 95 for SIU, and 89 for LEU\@. It contains 189 easy, 148 medium, and 86 hard questions. Fig.~\ref{fig:stat} presents the distributions by difficulty level and source dataset.

For each question, the model receives uniformly sampled RGB frames and the corresponding timestamp-aligned transcript. Frames for CAP are sampled from the annotated action interval, those for SIU from the interval associated with the target speaker, and those for LEU from the entire clip. All models are evaluated using the same questions, candidate options, transcripts, and sampled frame sets, subject to their input constraints. Because EgoVLM-360M accepts only a single frame~\cite{vinod2025egovlm}, we apply it independently to each sampled frame and aggregate the frame-level predictions through majority voting. In contrast, multi-frame-capable models process all sampled frames jointly.

A prediction is considered correct only if the extracted option letter exactly matches the ground-truth answer. Missing or unparsable responses are treated as incorrect. We report accuracy for the full benchmark and separately by question type and difficulty level. Because DeepSeek-V4-Flash-Vision-Exp and Qwen3.8-Max participate in determining the difficulty labels, we exclude their difficulty-stratified results to avoid circular evaluation. Their overall and task-specific results are retained.

\section{Results and Analysis}
\subsection{Comparison across Models}
% Table~\ref{tab:main_results} shows substantial differences across tasks and difficulty levels. Overall performance broadly increases with model scale, although this trend is not strictly monotonic across model families. Qwen3.8-Max and Claude-Sonnet-5 achieve the same overall accuracy of 77.54\%, but exhibit different capability profiles: Qwen3.8-Max performs better on scene understanding and GPT-5.6-Luna performs best on medium subset, whereas Claude-Sonnet-5 is considerably more robust on the hard subset. This indicates that overall accuracy alone cannot fully characterize a model’s strengths and weaknesses.
Table~\ref{tab:main_results} shows substantial variation across tasks and difficulty levels. Qwen3.8-Max and Claude-Sonnet-5 obtain the same overall accuracy of 77.54\%, but their task-specific results differ. Qwen3.8-Max performs better on LEU and SIU, reaching 94.38\% and 81.05\%, respectively, whereas Claude-Sonnet-5 obtains higher CAP accuracy at 71.55\%. Among the remaining models, Claude-Sonnet-5 records the highest accuracy on the easy and hard subsets, at 92.59\% and 46.51\%, while GPT-5.6-Luna obtains the highest medium-subset accuracy of 81.76\%. These results show that aggregate accuracy alone does not fully characterize a model's task-specific capabilities or its robustness to question difficulty.

Performance also varies with each model's temporal input capacity and training domain. EgoVLM-360M processes only one frame per inference and therefore cannot aggregate information over long video contexts, which likely contributes to its weak LEU performance. In contrast, egocentric-domain models perform strongly on SIU: EgoThinker-v1 achieves 63.16\%, outperforming Qwen3-VL-4B by 9.48 percentage points. This result is consistent with a benefit from egocentric adaptation for recognizing short-range speaker behaviors and intentions. However, the same advantage is less evident on CAP, which requires joint visual understanding, utterance comprehension, and action-sequence generation. This pattern suggests that domain adaptation does not consistently compensate for limited general visual-language reasoning.

Among the remaining models, most exhibit a marked performance decline from the easy subset to the medium and hard subsets. Closed-source models generally retain higher accuracy on the hard subset: Claude-Sonnet-5 achieves 46.51\%, followed by GPT-5.6-Luna at 40.70\% and Gemini-Flash-3.8 at 38.37\%. Among open-weight models, Qwen3-VL-4B-Instruct and EgoThinker-v1 achieve 37.21\% and 36.05\%, respectively. These results indicate that aggregate accuracy can mask substantial differences in model robustness as difficulty increases.
\subsection{Ablation Study}
\begin{table}[htbp]
\centering
\caption{
Effect of reasoning mode on Qwen3-VL series.
}
\label{tab:reasoning_ablation}
\setlength{\tabcolsep}{2.8pt}
\renewcommand{\arraystretch}{1.05}
\small
\resizebox{\columnwidth}{!}{
\begin{tabular}{lccccc}
\toprule
Setting
& Overall
& LEU
& SIU
& CAP
& \shortstack{Efficiency\\(samples/s)} \\
\midrule

\multicolumn{6}{l}{\textit{Qwen3-VL-2B}} \\

Instruct
& 39.48
& 42.70
& 22.11
& 45.19
& \textbf{1.60} \\

Thinking
& \textbf{59.10}\up{19.62\,pp}
& \textbf{58.43}\up{15.73\,pp}
& \textbf{54.74}\up{32.63\,pp}
& \textbf{61.09}\up{15.90\,pp}
& 0.030\down{53.6$\times$} \\

\midrule
\multicolumn{6}{l}{\textit{Qwen3-VL-4B}} \\

Instruct
& 56.03
& 68.54
& 53.68
& 52.30
& \textbf{1.51} \\

Thinking
& \textbf{73.05}\up{17.02\,pp}
& \textbf{80.90}\up{12.36\,pp}
& \textbf{66.32}\up{12.64\,pp}
& \textbf{72.80}\up{20.50\,pp}
& 0.075\down{20.2$\times$} \\

\bottomrule
\end{tabular}
}
\end{table}
\begin{table}[htbp]
\centering
\caption{Modality ablation study on CAP with different VLMs (\%).}
\label{tab:visual_ablation}
\setlength{\tabcolsep}{5pt}
\renewcommand{\arraystretch}{1.02}
\small
%\begin{tabular}{llc}
\begin{tabularx}{\linewidth}{
@{}
>{\raggedright\arraybackslash}p{3.5cm}
>{\raggedright\arraybackslash}p{2.5cm}
>{\centering\arraybackslash}X
@{}
}
\toprule
Model & Text only & Vision + Text \\
\midrule
Qwen3-VL-2B
& 43.93
& \textbf{45.19} \up{1.26\,pp} \\
InternVL2.5-2B
& \textbf{30.96}
& 26.78 \down{4.18\,pp} \\
EgoMind-7B 
& 39.75
& \textbf{48.54} \up{8.79\,pp} \\ 
\bottomrule
\end{tabularx}
\end{table}
We conduct two ablation studies to examine the effects of reasoning mode and visual input. First, we compare the Instruct and Thinking variants of Qwen3-VL-2B and Qwen3-VL-4B under identical input settings. As shown in Table~\ref{tab:reasoning_ablation}, Thinking improves the overall accuracy of Qwen3-VL-2B from 39.48\% to 59.10\%, an increase of 19.62 percentage points. The largest improvement occurs on SIU, where accuracy increases by 32.63 percentage points, while CAP improves by 15.90 percentage points. A similar trend is observed for Qwen3-VL-4B: its overall accuracy increases from 56.03\% to 73.05\%, with the largest task-specific increase of 20.50 percentage points on CAP\@. These consistent improvements across model scales and tasks demonstrate that extended reasoning benefits long-context understanding, speaker-intention inference, and embodied action planning.

These gains, however, introduce substantial computational overhead. For Qwen3-VL-2B, inference throughput decreases from 1.60 to 0.030 samples/s, making the Thinking variant approximately $53.6\times$ slower. For Qwen3-VL-4B, throughput decreases from 1.51 to 0.075 samples/s, corresponding to a slowdown of approximately $20.2\times$. Although long-form reasoning substantially improves accuracy, its reduced inference efficiency limits its applicability to online social agents that require timely responses. The Instruct variants therefore offer a more practical trade-off between accuracy and efficiency.

We further investigate the contribution of visual observations to CAP by comparing the text-only and vision--text settings across three VLMs. As shown in Table~\ref{tab:visual_ablation}, adding sampled frames improves Qwen3-VL-2B by 1.26 percentage points and EgoMind-7B by 8.79 percentage points, but reduces InternVL2.5-2B performance by 4.18 percentage points. The larger increase achieved by EgoMind-7B indicates that this model benefits more from visual input in the evaluated setting. Nevertheless, the inconsistent effects across models show that adding visual observations does not consistently improve CAP\@. These results indicate that the usefulness of visual context depends on the model's ability to integrate multi-frame visual and linguistic information.
\subsection{SocialCoAct-Agent Deployment}
\label{sec:robot_deployment}
We deployed \agentname{} on two RTX 5090 GPUs. Specifically, the agent integrates Whisper-Tiny for ASR, Qwen3-VL-8B for response and action planning, Qwen3-TTS for speech synthesis, and MuJoCo or a Unitree G1 for action execution. We evaluated each setting over ten runs using a 3.66\,s speech input and visual frames. All trials successfully produced both a verbal response and an executable gesture. The median end-to-end latency was 12.63\,s in simulation. The measured processing latency in the robot-deployment configuration was 7.57\,s, excluding camera transmission, robot communication, and physical execution.
% \section{Discussion and Limitations}
% \benchname{} evaluates social understanding and action planning but does not directly assess open-ended verbal response generation. CAP conditions action planning on a given verbal reply, enabling objective evaluation while leaving response generation outside the scope of the benchmark. Because multiple responses may be appropriate in the same context, single-reference metrics cannot reliably capture response quality. Future work could combine multiple references, structured evaluation criteria, and human judgments to evaluate verbal and bodily responses jointly.

% Another limitation concerns the granularity of the atomic-action vocabulary. The current action types have distinct communicative functions and affective meanings, which facilitates consistent annotation and unambiguous MCQs but may oversimplify action selection and overlook subtle behavioral variations. Future work could incorporate semantically similar actions, continuous motion parameters, and finer-grained action--utterance alignment to support more realistic and challenging evaluation.
\section{Conclusion}
In this work, we introduce \frameworkname{}, an agent-centered framework for egocentric social co-action with VLM-based humanoid robots. We have formulated egocentric social co-action as an end-to-end task and developed \agentname{}, which transforms egocentric observations and dialogue history into verbal responses and atomic actions aligned with corresponding utterance spans. The resulting actions are executable in MuJoCo and on a physical humanoid robot. To evaluate the VLM capabilities required by this task, we have constructed \benchname{}, an egocentric multimodal benchmark comprising video and timestamp-aligned transcripts from complex, multi-participant, and multi-scene interactions. Its three tasks, LEU, SIU, and CAP, assess long-context understanding, speaker-intention recognition, and coordinated action planning, respectively. In particular, CAP represents embodied responses as human-interpretable atomic actions explicitly aligned with given utterance spans.

Our experiments show that performance varies across model scales, temporal input capacities, and training domains, while CAP and hard social-interaction questions remain particularly challenging. Reasoning ablations demonstrate substantial trade-offs between accuracy and efficiency, while modality ablations show that the benefits of visual context vary across models. By connecting multimodal social understanding with interpretable, utterance-aligned action planning and execution, \frameworkname{} provides a unified foundation for evaluating and developing socially responsive VLM-based humanoid agents.

{\small
\bibliographystyle{IEEEtran}
\bibliography{refs}

\end{document}

\end{document}